# A Selective Temporal Hamming distance to find patterns in state transition event timeseries, at scale

Sylvain Marié [1] and Pablo Knecht [1]

[1] AI Hub, Schneider Electric, IntenCity, 160, Avenue des Martyrs, 38000 Grenoble, France
E-mail: sylvain.marie@se.com, pablo.knecht@se.com
ORCID: *0000-0002-5929-1047*, *0009-0008-3361-1832*

**Summary:** Discrete event systems are present both in observations of nature, socio economical sciences, and industrial systems. Standard analysis approaches do not usually exploit their dual event / state nature: signals are either modeled as transition event sequences, emphasizing event order alignment, or as categorical or ordinal state timeseries, usually resampled – a distorting and costly operation as the observation period and number of events grow. In this work we define state transition event timeseries (STE-ts) and propose a new Selective Temporal Hamming distance (STH) leveraging both transition time and duration-in-state, avoiding costly and distorting resampling on large databases. STH generalizes both resampled Hamming and Jaccard metrics with better precision and computation time, and an ability to focus on multiple states of interest. We validate these benefits on simulated and real-world datasets.

## 1. Introduction

Discrete state systems (DSS) are systems with a discrete set of states (a.k.a. a discrete *state space*). They are present in numerous domains from robotics to industrial processes, energy, mobility, social sciences, games or biology [1-8]. Discrete event systems (DES) are DSS where state transition is based on instantaneous transition events [9]. Modeling of DES has been studied for over 35 years, typically with Finite State Automata, Petri Nets, Vector DES, Event Graphs, Queuing systems, Markov Processes [5]. Techniques such as Hidden Markov models are also used to estimate hidden states and transitions [8].

Data collected from DES' observation can be represented either as a sequence of state transition events, or as a timeseries of categorical states. Two families of approaches in the literature are therefore relevant: analysis of *event sequences*, and analysis of *categorical timeseries*.

In many statistical and machine learning methods, defining a proper similarity or distance between samples plays an important role: e.g. in *clustering* with Agglomerative clustering, Spectral clustering, DBScan, etc. ; in *classification* with KNN, SVM, etc.; in *visualization* with Isomap, MDS [10]. A known difficulty with categorical variables as opposed to real-valued is the absence of observed magnitude of difference. Metrics include association measures, Kramer's v, Kendall Tau, simple matching (SM), (inverse) Occurrence Frequency, Elkin, Goodall, LIN and derivatives, Total variation distance, Kullback-Leibler divergence. A common practice is to binarize data with presence/absence indicators and use binary metrics, as in the Jaccard index, or in [11,12]. Information theory provides interesting similarities [12,13]. See [14] for a review, generalized in [10].

Analysis of categorical *event sequences* refers to counts of common states, and edit distances such as Levenshtein, Longest Common Subsequence, Hamming, Spectrum kernel, MCA, Optimal Matching (OM) including MSW and BLAST [15-18]. Correlated sequences are also found with association rules mining techniques including fuzzy sets, FP-Growth or Apriori algorithm [19,20]. State-aware metrics weight events according to specific state transitions, such as Dynamic Hamming Distance or OM for transitions [21, 15]. See [15] and [7] for a status of the field.

Analysis of *categorical timeseries* (CTS) refers to $\chi^2$, Hamming, fuzzy comparison, Cramer's v, possibly extended to support lags, and Pearson correlation applied to binarized representations [16, 22]. The case of binary series is worth mentioning with metrics such as Pearson's Phi, Simple Matching coefficient (SMC) (a.k.a Rand), Hamming, Jaccard Index [23], with lag tolerance [18, 24], or time-depending weigths [25]. Finally, methods exist for *real-valued* timeseries such as cross-correlation, DTW and its variants [26], metric learning [27], and tools such as Matrix Profile [28]. Time warp edit distance [29] is an interesting step in the direction of applying them to categorical timeseries.

## 2. Challenge, related work and constrictions

Most metrics require timeseries to be uniformly sampled. Yet, resampling non-uniform timeseries is a computationally intensive task inducing distortion [30]. To the best of our knowledge, only two recently proposed metrics overcome this critical issue. FTH [31] is an edit distance accounting for time shifts required in mobility data analysis, thanks to fuzzification. It however lacks native symmetry, is a semi-metric, requires Fast Fourier Transform, and has a complexity of $\mathcal{O}(max(n,m)^2 \log max(n,m))$ (the complexity in [31] seems incorrect as NFFT's complexity varies with $n$ not $T$ [32]), reducing its applicability to large series. While a draft of temporal Hamming concept is mentioned as a singular case, it is not formalized nor studied, and its complexity analysis $\mathcal{O}(T)$ seems incorrect (see section 4.1). Temporal Similarity [33] is a temporal version of Jaccard computed online in linear time. It however restricts to binary series, does not address the situation where both series are 0, and does not prove metric properties.

**Contributions** In this work we suggest a new formal definition of *State Transition Event timeseries* (STE-ts) and propose a *Selective Temporal Hamming* (STH) similarity and associated distance. STH has the following benefits:
- It is a formal generalization of the Hamming and Jaccard metrics for continuous time.
- It is equivalent to metrics computed on infinitely small sampling periods, avoiding any distortion.
- It generalizes the Jaccard metric to non-binary series, including ambiguous states ("excluded").
- It is a proper metric (incl. triangular inequality) in many cases and can thus be easily used in a kernel.
- Its linear complexity $\mathcal{O}(n+m)$ solely depends on the number of events, as opposed to resampled Hamming and Jaccard $\mathcal{O}(n+m+n^{(P)})$ that also depend on the number $n^{(P)}$ of sampling buckets.

The rest of this paper is organized as follows: in Section 3 we remind DES, define state transition event timeseries (STE-ts) and remind existing metrics for uniformly sampled STE-ts. In Section 4 we introduce novel distance metrics and detail a few interesting properties. In Section 5 we present experiments on simulated and real-world datasets, highlighting the benefits. Finally, we conclude in section 6.

## 3. State Transition Event Timeseries

### 3.1. Discrete Event Systems

We consider a discrete event system **DES** with set of states $\mathcal{S}=\{\sigma_k\}$ and set $\mathcal{T}$ of possible state transitions defined as a set of pairs (before state, after state): $\mathcal{T} = \{tr_i = (b_i, a_i)\} \subset \mathcal{S} \times \mathcal{S}$. In the following work we focus on *state-changing transitions*, i.e. $b_i \neq a_i \forall i$. Note that any system whose state is represented by a finite number of categorical variables can be represented as such. This includes Markov Processes (a.k.a. Markov Chains) and HMMs. **Fig. 1** illustrates two prototypic DES: a simplified DVD player system and an industrial alarm system with shelving capability. Note: *shelving* (a.k.a. *muting*) means that the true alarm state is unknown and not relevant according to the user, for example during a maintenance operation.

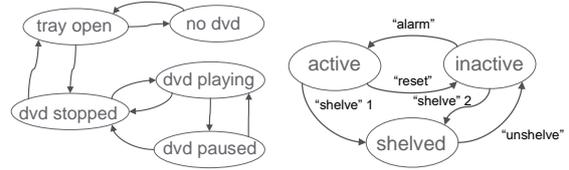

**Fig. 1.** Simple DVD Player (left) ; Alarm system (right).

**Property:** (*Simplification*) merging some states and associated transitions. e.g., merging "dvd playing" and "dvd paused", leads to valid DES representations.

### 3.2. State transition event sequences and timeseries

We define a *sequence* of state transition events **STE-seq** of length $n$ as an ordered collection of $n$ state-changing transitions $\{tr_k = (b_k, a_k)\}_{k=1,...,n} \subset \mathcal{T}^n$, with state-changing constraint $\{b_k \neq a_k\} \forall k$ and sequence consistency constraint $\{a_k = b_{k+1}\} \forall k < n$. Such a sequence can be rewritten as a *categorical sequence* of $n+1$ states $\{\sigma_k\}_{k=0...,n} \subset \mathcal{S}^{n+1}$ with state changing constraint $\{\sigma_k \neq \sigma_{k+1}\} \forall k < n$.

Let $\Omega$ be the set of all timestamps. We define a *timeseries* of state transition events **STE-ts** of length $n$, as an ordered collection of $n$ timestamped state-changing transitions $\{(t_k, tr_k)\}_{k=1,...,n} \subset (\Omega \times \mathcal{T})^n$ complying with the above constraints, associated with a *start time* $t_0$ and an *end time* $t_{n+1}$. Such a timeseries can be rewritten as a *categorical timeseries* with $n+1$ pairs of state value and timestamp $\{(t_k, \sigma_k)\}_{k=0,...,n} \subset (\Omega \times \mathcal{S})^{n+1}$, associated with an end time. A given state $\sigma_k$ is valid for the interval $[t_k, t_{k+1}[$.

Note that the above definition is similar to the traditional definition of *categorical timeseries* (CTS) [16], with three key restrictions: (a) a state value is valid for the interval following the timestamp, until next timestamp ; (b) there is an explicit end time; and (c) all values differ from previous.

### 3.3. Prior art – reference resampled metrics

We now consider two STE-ts $s_i$ and $s_j$ with $n$ and $m$ transition events respectively, with the same start and end time and total duration $T$. We remind below the two simplest, most commonly used metrics, defined on resampled series. Resampled series $s_i^{(P)}, s_j^{(P)}$ are derived with sampling period $P$, sampling rate $F = 1/P$, resulting in (finite) number of samples $n^{(P)} = T/P$. We note $\sigma_{ik}^{(P)}$ and $\sigma_{jk}^{(P)}$ the respective resampled values $\forall k \in 1..n^{(P)}$. The *Hamming* distance $HD$ is obtained as a count of non-matching samples:

$$HD(s_i^{(P)}, s_j^{(P)}) = |k = 1..n^{(P)} \mid \sigma_{ik}^{(P)} \neq \sigma_{jk}^{(P)}| \quad (1)$$

The *normalized Hamming* distance $nHD$ is defined as $nHD = HD/n^{(P)}$. It represents the approximate proportion of the time during which the two series' values differ. The quality of the approximation globally increases with the sampling rate but is subject to local distortion (see section 5.1). $HD$ and $nHD$ are proper metrics complying with *positivity*, *symmetry*, *distinguishability*, and *triangular inequality* [34]. We note $H=n-HD$; $nH=1-nHD$ the associated similarities.

The Jaccard similarity index $J$ is defined for a binary representation of feature presence/absence (2) on static observations [23]. It is also used in practice on resampled series as a biased alternative to Hamming where value "0" has less interest than "1" [18, 24].

$$J(s_i^{(P)}, s_j^{(P)}) = \frac{|k| \sigma_{ik}^{(P)} = \sigma_{jk}^{(P)} = 1|}{|k| \sigma_{ik}^{(P)} = \sigma_{jk}^{(P)} = 1| + |k| \sigma_{ik}^{(P)} = 1, \sigma_{jk}^{(P)} = 0| + |k| \sigma_{ik}^{(P)} = 0, \sigma_{jk}^{(P)} = 1|} \quad (2)$$

The Jaccard distance is defined as $JD = 1 - J$ with $J(\mathbf{0},\mathbf{0}) := 1$. It meets *positivity*, *symmetry*, *distinguishability*, and *triangular inequality* [23, 35].

**Property**: (*Linear complexity*) Assuming initial ordering, resampling of a single timeseries $s_i$ is $\mathcal{O}(n^{(P)} + n)$ as for each bucket there is a need to find the events that fall in this bucket. Therefore *H, HD, nH, nHD, J,* and *JD* have complexity $\mathcal{O}(n^{(P)} + n + m)$.

<u>Note</u>: in this paper we use the simplified notations (e.g. $HD$, $JD$) for all resampled metrics except for places where it matters to remind the $P$: $HD^{(P)}, JD^{(P)}$.

## 4. Temporal metrics for STE-ts

We now consider the collection of time intervals $\mathcal{I}_{ij} = \{\iota_k = [t_k, t_{k+1}[\}$ defined by the union of all timestamps in $s_i, s_j$. Its cardinal depends on the number of timestamps that are common to $s_i$ and $s_j$: $\max(n,m) + 2 \leq |\mathcal{I}_{ij}| \leq n + m + 2$. For each interval $\iota$ we define $\Delta_\iota$ its duration, and $\sigma_i^\iota$ and $\sigma_j^\iota$ the respective states of $s_i$ and $s_j$ on $\iota$. Note that $T = \sum_{\iota \in \mathcal{I}_{ij}} \Delta_\iota$.

### 4.1. Temporal Hamming similarity and distance

We define the *Temporal Hamming* Similarity *TH* as the sum of durations of all intervals in $\mathcal{I}_{ij}$ where the two series $s_i, s_j$ have the same value (3).

$$TH(s_i, s_j) = \sum_{\iota \in \mathcal{I}_{ij} \mid \sigma_i^\iota = \sigma_j^\iota} \Delta_\iota \quad (3)$$

The *Normalized Temporal Hamming* similarity *nTH* is obtained by dividing *TH* by the total duration:

$$nTH(s_i, s_j) = \frac{TH(s_i, s_j)}{T} = \frac{\sum_{\iota \in \mathcal{I}_{ij} \mid \sigma_i^\iota = \sigma_j^\iota} \Delta_\iota}{\sum_{\iota \in \mathcal{I}_{ij}} \Delta_\iota} \quad (4)$$

We define for *TH* and *nTH* their associated distances $THD = T - TH$ and $nTHD = 1 - nTH$.

**Property:** (*Metric*) *THD* and *nTHD* are proper *distance* metrics. *TH* and *nTH* are *similarity* measures.

**Property**: (*nH equivalence*) when all events have uniform timestamps, all interval durations are identical. Temporal Hamming metrics equal 'resampled' Hamming: $nTH = nH$, $nTHD = nHD$.

**Property**: (*Limit*) $nTH$ (resp. $nTHD$) is the limit of $nH^{(P)}$ (resp. $nHD^{(P)}$) as sampling period $P$ approaches zero:

$$nTH[D](s_i, s_j) = \lim_{P \to 0} nH[D](s_i^{(P)}, s_j^{(P)}) \quad (5)$$

We similarly define Temporal Jaccard $TJ$ for binary series (6), associated distance $TJD = 1 - TJ$, and note that analogous properties ($J$ equivalence, sampling limit) hold.

$$TJ(s_i, s_j) = \frac{\sum_{\iota \in \mathcal{I}_{ij} \mid \sigma_i^\iota = \sigma_j^\iota = 1} \Delta_\iota}{T - \sum_{\iota \in \mathcal{I}_{ij} \mid \sigma_i^\iota = \sigma_j^\iota = 0} \Delta_\iota} \quad (6)$$

**Property**: (*T-S equivalence*) Temporal Jaccard $TJ$ equals the *temporal similarity* as defined in [33] for single alarm sets $A = \{s_i\}, B = \{s_j\}$.

### 4.2. Selective Temporal Hamming metric

We now consider a state partition: $S = S_I \cup S_O \cup S_E$ with
- $S_I$ a set of states of *interest*. For example "active", or {"dvd playing", "dvd paused"}. We define $sim$ a similarity function between states in $S_I$ as in [31]
- $S_O$ a set of *other* states, e.g. "inactive", "dvd stopped"
- $S_E$ a set of *excluded* (or "ambiguous") states. For example "shelved", or {"tray open", "no dvd"}.

We define the *Selective Temporal Hamming* (STH) similarity for two STE-ts $s_i$ and $s_j$, for state sets $\{S_I, S_O\}$, as in (6):

$$STH_{\{S_I, S_O\}}(s_i, s_j) = \begin{cases} ?\ (undef) & \text{if } \forall \iota,\ \sigma_i^\iota \in S_E \text{ or } \sigma_j^\iota \in S_E \\ 1 & \text{if } \forall \iota,\ \sigma_i^\iota \notin S_I \text{ and } \sigma_j^\iota \notin S_I \\ x(s_i, s_j) & \text{otherwise} \end{cases} \quad (6)$$

with

$$x(s_i, s_j) = \frac{\sum_{\iota \in \mathcal{I} \mid \sigma_i^\iota, \sigma_j^\iota \in S_I^2} sim(\sigma_i^\iota, \sigma_j^\iota) \cdot \Delta_\iota}{\sum_{\iota \in \mathcal{I} \mid \sigma_i^\iota, \sigma_j^\iota \in (S_I^2 \cup (S_I \times S_O) \cup (S_O \times S_I))} \Delta_\iota} \quad (7)$$

In this paper we restrict our analysis to $sim$ being the identity function: 1 when $\sigma_i^\iota = \sigma_j^\iota$ and 0 otherwise. The *STH* similarity can be interpreted as the ratio between the time during which both series have the same state *of interest* and the time during which at least one of the series has a state of interest. The *excluded* state discards intervals. We define the Selective Temporal Hamming *distance* as $STHD = 1 - STH$.

**Property**: (*Normalized*) STH values are in [0,1]. STH=1 means that on all intervals, either both series have a state of interest and it is the same ($\sigma_i^\iota = \sigma_j^\iota$), or none of them has a state of interest. STH=0 means that there is no interval during which $s_i$ and $s_j$ have the same value and this value is of interest.

**Property**: (*Definition*) STH is *undefined* when for each interval $\iota \in \mathcal{I}$, at least one series has a state in $S_E$. This situation is similar to the one caused by missing data in real-valued timeseries. A good fallback value in this case is application dependent. The authors suggest zero (0) as default.

**Property**: (*Hamming equivalence*) STH with all states "*of interest*" ($S_I = S$) is the *normalized Temporal Hamming nTH*. As such, all properties of $nTH$ hold.

**Property**: (*Jaccard equivalence*) for binary STE-ts, STH with $S_I = \{1\}, S_O = \{0\}, S_E = \emptyset$ is the *Temporal Jaccard TJ*. As such, all properties of $TJ$ hold.

**Property**: (*Metric*) When $S_E = \emptyset$ and $|S_O| \leq 1$, the STHD distance satisfies both the *positivity*, *symmetry*, *distinguishability*, and *triangular inequality*. $|S_O| \leq 1$ can be relaxed if *distinguishability* is not needed. A proof is provided in [36].

**Property**: (*Complexity*) Both *[n]TH, TJ* and *STH* have a linear complexity with respect to the number of intervals $|\mathcal{I}_{ij}|$ induced by $s_i$ and $s_j$, so are $\mathcal{O}(n + m)$.

**Table 1** summarizes the various metrics and their properties, and the two state-of-the-art metrics without resampling, FTH [31] and TS [33], as reference.

## 5. Experiments

### 5.1. Execution time and precision benchmarks

In this section we perform controlled experiments with simulated STE-ts to validate the two main benefits of *STH* over resampled metrics such as Hamming or Jaccard: execution time and precision. For value comparison purposes we use the Hamming configuration ($S_I = S$) for *STH*, so *STH = nTH* in this section and its values are compared to the resampled normalized Hamming $nH^{(P)}$. Yet, similar results can be observed by choosing a Jaccard configuration ($S_I = 1, S_O = 0$) and comparing *STH(=TJ)* to resampled $J^{(P)}$.

Results are obtained on a PC with an AMD Ryzen 5 pro 5650U 2.3Ghz processor and 16Gb ram. Resampling is done with the `pandas` library [37,38].

**Execution time vs. number of events**

We create *Rts(n)* a timeseries with *n* randomly occurring binary state change events spanning 30 days.

We also create *Tk*, a timeseries spanning 30 days with 8640 evenly spaced (/5mins) events, with added random uniform time lags ranging from 1ms to 200ms.

**Fig. 2** (left) shows average computation times (over 30 runs) of *STH* and $nH^{(P)}$ with 5 min resampling, for pairs *Rts(n)* vs. *Rts(n)*, *Rts(n)* vs. *Rts(1)* and *Rts(n)* vs. *Tk*. Both metrics execution time grow linearly with the number of events in the series, confirming our complexity analysis. **Fig. 2** (right) shows the execution time ratio $t(nH^{(P)})/t(STH)$. *STH* is between 3.5-14 times faster than $nH^{(P)}$ depending on the series. The largest improvements [7.5x-14x] are obtained when one of the series has only one event (*Rts(1)*).

**Execution time vs resampling period**

We now generate two periodic binary STE-ts spanning 1 month (**Fig. 3**): $PS_0$ with 14min period and 60% duty cycle; and $PS_{1/3}$ obtained by adding to $PS_0$ a lag of 1/3 of a period.

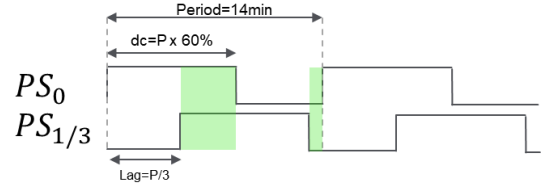

**Fig. 3.** two periodic STE-ts with matching states highlighted.

We measure the average computation time over 30 runs of $nH(PS_0, PS_{1/3})$, for a wide range of resampling periods, and compare it with that of *STH=nTH*. Results in **Fig. 4** (left) confirm that $nH^{(P)}$ is $\mathcal{O}(T/P)$: as the sampling period *P* decreases, it is up to 4950x slower.

**Metric precision**

By design, $PS_0$ and $PS_{1/3}$ have the same state 1/3 of the time. Therefore, their normalized Hamming distance should be 2/3 if measured on an infinite number of periods. *STHD=nTHD* is computed and is 0.6666574 as expected. We compute $nHD^{(P)}$ for various sampling periods *P* and compare it with *STHD*. **Fig. 4** (right) shows the impact of resampling distortion on $nHD^{(P)}$. For some sampling periods the series are even considered identical by *nHD* (*nHD*=0)! *STHD* always guarantees the most accurate value.

**Table 1** summarizes the results of this section.

### 5.2. Experiments on new analysis capabilities

In this section we illustrate the capability of *STH* to generalize *Jaccard* to non-binary series. We will use Clustering to highlight the impact of various choices of states to include in $S_I$ and $S_E$. For two public datasets below, raw data is first transformed to event transitions timeseries. We then compute a pairwise distance matrix for each pair of STE-ts, for various distance metrics. We finally apply agglomerative clustering on resulting distance matrices and comment.

**Table 1.** Metrics properties, along with speed and precision experiments summary. '*' means '*under conditions*'.

| method | binary | categ. | metric | complexity | speed | distortion |
|---|---|---|---|---|---|---|
| (resampled) Jaccard | x | | x | $\mathcal{O}(n+m+\frac{T}{P})$ | baseline | 0 − 100% (Fig.4 right) |
| (resampled) Hamming | x | x | x | | | |
| Selective Temporal Hamming $STH$ | x | x | * | $\mathcal{O}(n+m)$ | 3.5 − 4950× faster (Fig.2, Fig.4 left) | None |
| Temporal Hamming $= STH_{S,\emptyset}$ | x | x | x | | | |
| Temporal Jaccard $= STH_{\{1\},\{0\}} = TS$ | x | | x | | | |
| Fuzzy Temporal Hamming $FTH$ | x | x | ? | $\mathcal{O}(max(n,m)^2 log(max(n,m)))$ | N/A | N/A |

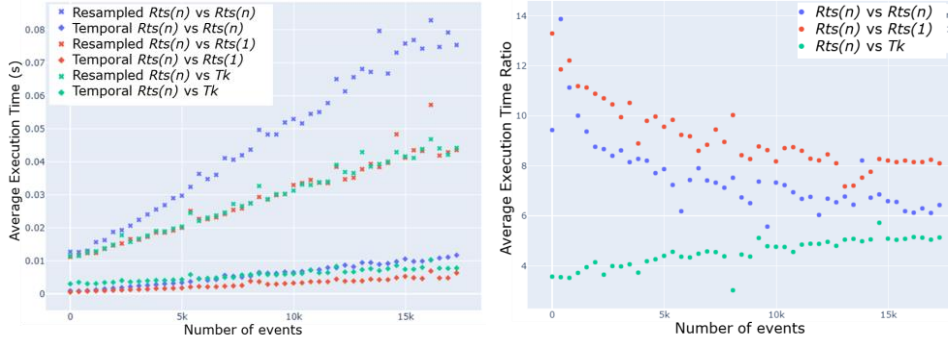

**Fig. 2.** Execution time (left) and ratio (right) for Resampled vs. Temporal Hamming distances for various pairs of series.

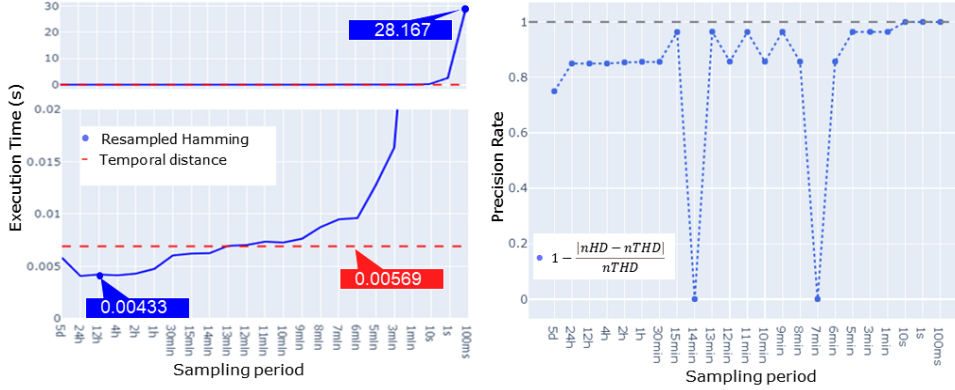

**Fig. 4.** Left: execution time (s) for resampled Hamming distance (blue) vs. STH=nTH (dashed red), for various sampling periods (x-axis). Bottom left: zoom on y-axis. Right: precision rate of *nHD* (1=no distortion) for various sampling periods.

**Leveraging $\mathcal{S}_I$ to focus on specific states**

We can focus *STH* on specific states of interest $\mathcal{S}_I$, the same way we use Jaccard to focus on the "1" in binary series. We illustrate this on the *US Weather Events* dataset, employed in a study to discover propagation and influential patterns [39]. It contains a collection of weather events data across 2071 US locations. Possible states are: *Cold, Fog, Snow, Hail, Rain, Storm, Precipitation,* and *Normal*, which represents the absence of notable weather event. Our analysis focuses on 2019, for 1000 random locations.

We run clustering with 3 settings of *STH*. The number of clusters *k* is selected from the dendrogram based on decreasing the optimal number found by Silhouette score until an acceptable macroscopic geographical view is found. Settings are:

a) $\mathcal{S}_I = \mathcal{S}$ so *STH=nTH* (Hamming) ; *k*=30
b) $\mathcal{S}_I = \mathcal{S}\setminus\{Normal\}, \mathcal{S}_O = \{Normal\}$; *k*=30
c) $\mathcal{S}_I = \{Snow, Hail\}, \mathcal{S}_O = \mathcal{S}\setminus\{Snow, Hail\}$ ; *k*=30

Results are shown in **Fig. 5**, where each marker in the map represents a location, and its color and shape represent the cluster id. Whereas Hamming (a) reveals a large-size 'blue' cluster, in (b) since the *Normal* (no event) state is removed from $\mathcal{S}_I$, differences between abnormal states are highlighted, for example on the US east coast; in (c) only snow and hail are in $\mathcal{S}_I$, highlighting differences in north-eastern regions but not so much in southern ones anymore.

**Leveraging $\mathcal{S}_E$ to handle ambiguity**

The new set $\mathcal{S}_E$ can be used to deal with ambiguous states where it is preferable to not derive any similarity value. We illustrate this on the *MASS SS3 Sleep Annotations* dataset, that contains sleep stages annotations for an entire sleep period of 62 patients [40]. The states are *{?, W,1,2,3,R}*. R (rapid eyes movement) is the state of interest, while W (awake), and the three levels of sleep depth (1, 2, 3) are not. The absence of annotation on an interval is replaced with a new *Sleep stage "E"* state.

We compare two STH settings:

a) $\mathcal{S}_I = \{R\}$ and $\mathcal{S}_E = \emptyset$ so *STH* is equivalent to *Jaccard* on a preprocessed dataset where *R* is mapped to 1 and all other states are mapped to 0
b) $\mathcal{S}_I = \{R\}, \mathcal{S}_O = \{W, 1,2,3\}$ and $\mathcal{S}_E = \{?, E\}$.

Resulting similarities are then used in the agglomerative clustering algorithm, with a fixed number of clusters (8). The patients' STE-ts are displayed in the heatmap of **Fig. 6** in the agglomeration order. Clusters are delimited by white stripes. We observe that clusters in (b) seem more pure in terms of content, and that some individuals with many missing or ambiguous values are better grouped. For example the individual identified by the arrow was isolated in a single cluster with Jaccard, and is now in a consistent group.

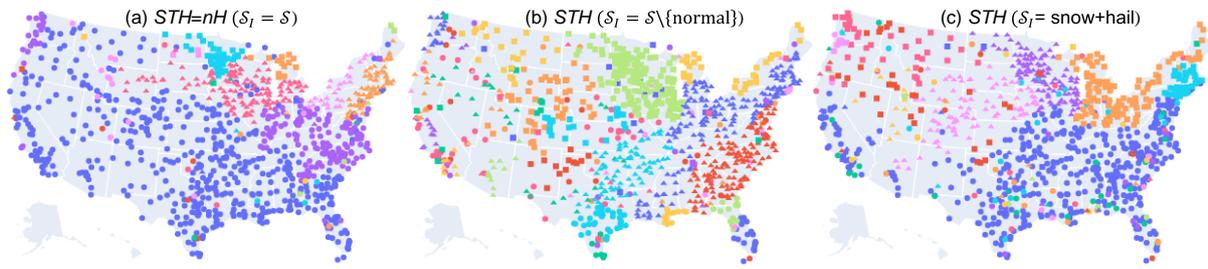

**Fig. 5.** US map with one marker per location, STH-based clusters identified by color and shape, for 3 sets of states of *interest*.

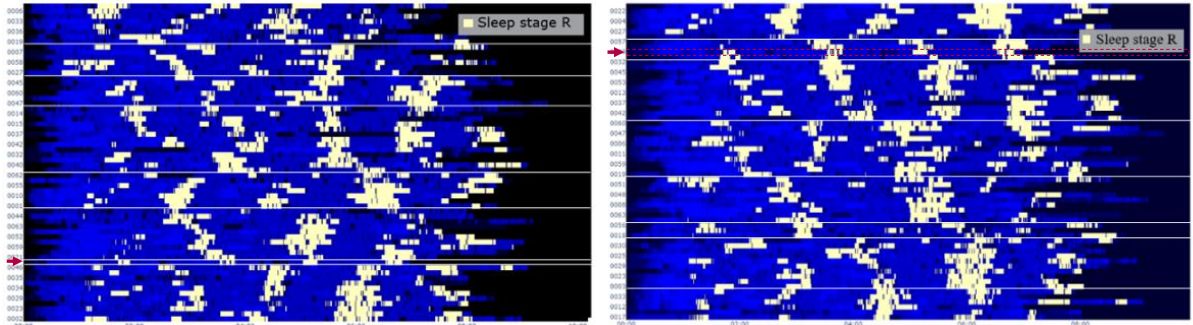

**Fig. 6.** Patients' (y axis) sleep state across time (x axis) sorted by aggregation order with 8 clusters (white stripes). Left: Jaccard $\{R$ vs. all$\}$; right: $\mathcal{S}_I = \{R\}$, $\mathcal{S}_E\{?, E\}$ (both black) $\mathcal{S}_O = \{W, 1, 2, 3\}$ (four shades of blue).

## 6. Conclusion and future work

In this paper we formalized state transition event timeseries (STE-ts), bridging *event sequences* and *categorical timeseries* formalisms in the context of Discrete Event Systems. We introduced the *Selective Temporal Hamming* metric, generalizing *Hamming* and *Jaccard* for continuous time. As opposed to these resampled metrics where a tradeoff between speed and precision is required, our experiments on simulated datasets confirm that *STH* avoids distortion and is faster to compute, making it particularly suitable for large scale data analysis. Moreover, *STH* also brings *Jaccard*-like capabilities for non-binary series. Our clustering experiments on real world datasets highlighted the ability to focus on states of interest, and to handle ambiguous states – both key features to inject subject matter expertise in the analysis. These properties make *STH* particularly well suited to compare STE-ts while not limiting its applicability to any categorical timeseries.

In the future we plan to study how *STH* can be used in a kernel, e.g. for SVM classification tasks. Also, in this paper we restricted the state similarity weights *sim* to the identity matrix; it would be interesting to study the conditions on *sim* under which metric properties of *STH* still hold. Finally, finding a way to tolerate time shifts without impacting complexity significantly is another interesting challenge.